\documentclass[10pt,twocolumn,letterpaper]{article}

\usepackage{iccv}              % To produce the CAMERA-READY version
\definecolor{iccvblue}{rgb}{0.21,0.49,0.74}
\usepackage[pagebackref,breaklinks,colorlinks,allcolors=iccvblue]{hyperref}

\def\paperID{5439} % *** Enter the Paper ID here
\def\confName{ICCV}
\def\confYear{2025}

\title{SOYO: A Tuning-Free Approach for Video Style Morphing via Style-Adaptive Interpolation in Diffusion Models}

\author{Haoyu Zheng$^1$ \quad Qifan Yu$^1$ \quad Binghe Yu$^2$ \quad Yang Dai$^1$ \\
Wenqiao Zhang$^1$ \quad Juncheng Li$^1$ \quad Siliang Tang$^1$ \quad Yueting Zhuang$^1$\\
\small{$^1$Zhejiang University, $^2$University of Electronic Science and Technology of China}\\
\small{$^1$\{\texttt{zhenghaoyu, yuqifan, daiyang, wenqiaozhang, lijuncheng, siliang, yzhuang}\}@zju.edu.cn,} \\
\small{$^2$\texttt{2021090923002@std.uestc.edu.cn}}
}

\begin{document}
\newcommand*{\method}{SOYO}

\twocolumn[{
\renewcommand\twocolumn[1][]{#1}%
\maketitle
\begin{center}
    \vspace{-1em}  
    \centering
    \captionsetup{type=figure}
    \includegraphics[width=\textwidth]{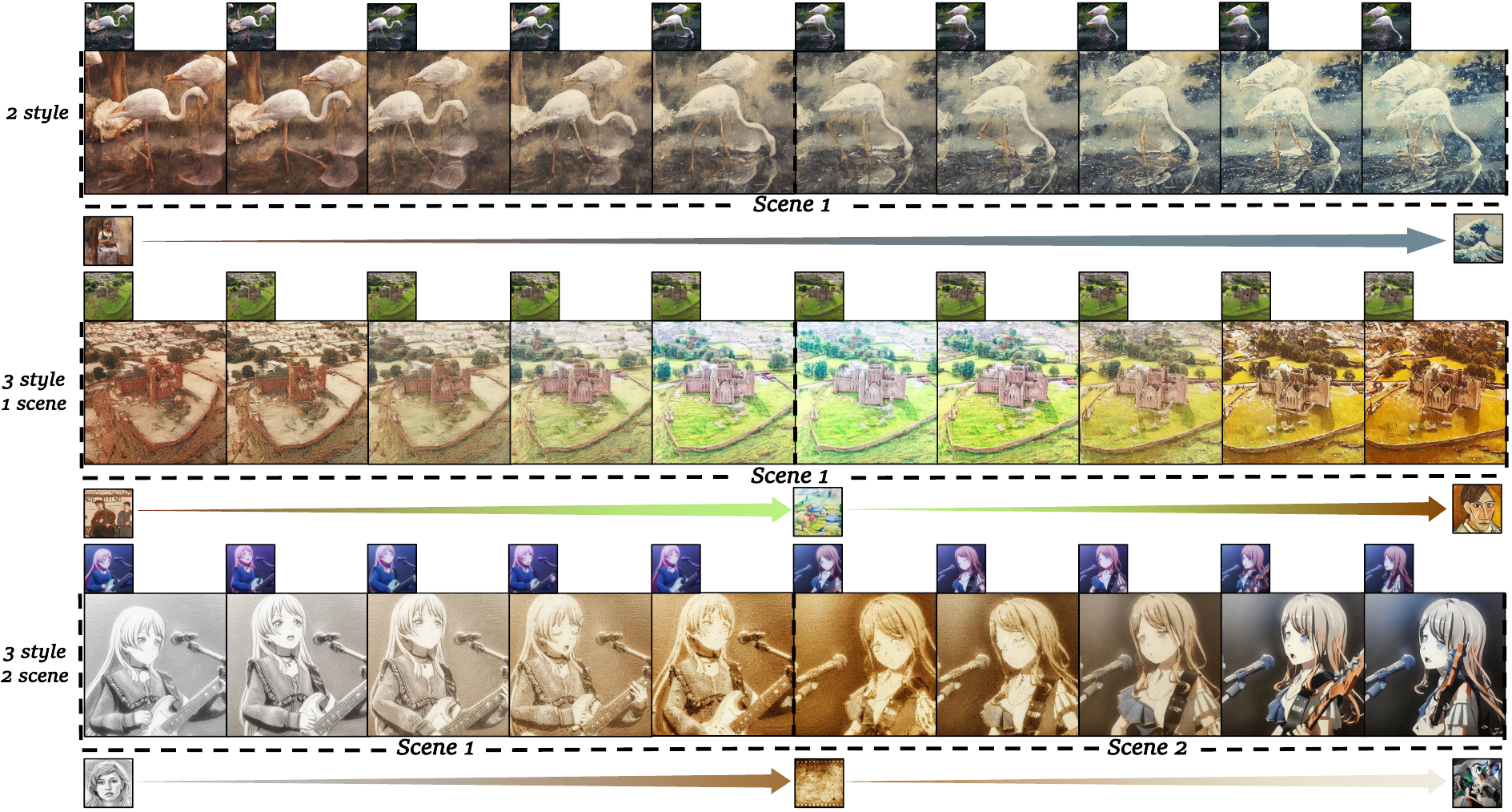}
    \captionof{figure}{Video style morphing results generated by SOYO.}
    \label{fig:title}

\end{center}
}]

\begin{abstract}
Diffusion models have achieved remarkable progress in image and video stylization. However, most existing methods focus on single-style transfer, while video stylization involving multiple styles necessitates seamless transitions between them. We refer to this smooth style transition between video frames as video style morphing. Current approaches often generate stylized video frames with discontinuous structures and abrupt style changes when handling such transitions. To address these limitations, we introduce SOYO, a novel diffusion-based framework for video style morphing. Our method employs a pre-trained text-to-image diffusion model without fine-tuning, combining attention injection and AdaIN to preserve structural consistency and enable smooth style transitions across video frames. Moreover, we notice that applying linear equidistant interpolation directly induces imbalanced style morphing. To harmonize across video frames, we propose a novel adaptive sampling scheduler operating between two style images. Extensive experiments demonstrate that SOYO outperforms existing methods in open-domain video style morphing, better preserving the structural coherence of video frames while achieving stable and smooth style transitions.
\end{abstract}    
\section{Introduction}
\label{sec:intro}

The rise of diffusion models has significantly advanced the fields of image and video generation~\cite{rombach2022high, ho2022classifier, blattmann2023stable,zheng2024laser, lin2025healthgpt}. At the same time, these models have also been applied to the image and video editing tasks~\cite{hertz2022prompt, cao2023masactrl, tumanyan2023plug, mokady2023null}. Among these applications, style transfer has garnered considerable attention~\cite{jeong2023training, wang2023stylediffusion, chung2024style}. Specifically, given a reference style image, the model modifies the content image or video to imbue it with the desired style. Existing video style transfer methods~\cite{ku2024anyv2v, huang2024style} based on diffusion models are typically limited to statically applying a single style across the entire video sequence. While this approach achieves a certain level of stylization, it lacks support for dynamic transitions between multiple styles. Such transitions are essential for conveying nuanced emotional tones, evolving atmospheres, and thematic shifts within the narrative. Direct application of existing methods, without dynamic style morphing, often results in abrupt or incoherent style changes, disrupting the visual flow and diminishing the narrative coherence. This limitation significantly restricts the flexibility of video style transfer in expressing complex moods and maintaining consistent emotional or thematic progression.

\begin{figure}[!t]
    \centering
    \includegraphics[width=1\linewidth]{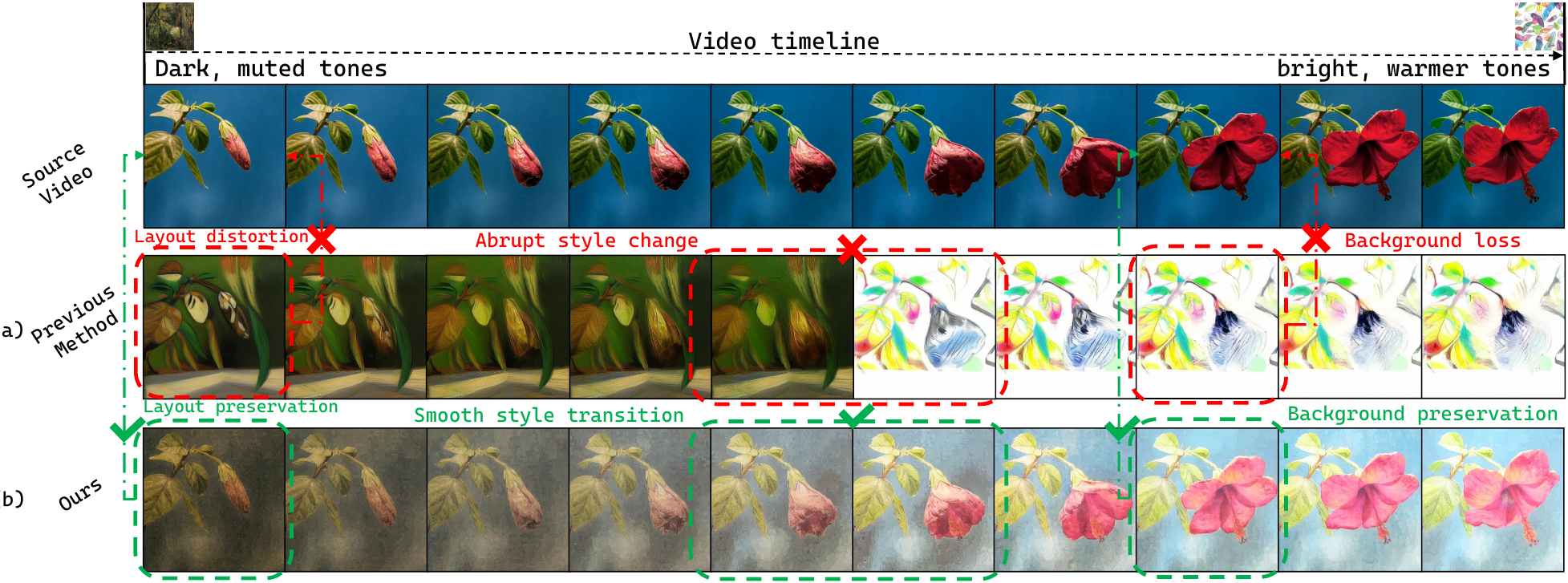} 
    \caption{In the context of video stylization tasks involving style transitions, abrupt style switching can result in visually disjointed effects. In contrast, a smooth and gradual style transition facilitates a visually natural transformation between styles.} 
    \label{fig:intro_case} 
\end{figure}

In contrast, style morphing offers rich visual coherence while enabling creative possibilities for artistic video production~\cite{jing2019neural, singh2021neural}. This capability serves artistic contexts requiring visual continuity and emotional resonance~\cite{block2020visual}. In cinema and animation, it visualizes character development through evolving aesthetics~\cite{bordwell2004film}, while in digital art, it enables integration of diverse visual languages—from photorealism to abstraction—creating nuanced compositions~\cite{elgammal2017can}. These transitions facilitate visual metaphors and contrasting aesthetics that evoke specific emotional responses. As Fig.~\ref{fig:intro_case} illustrates, gradual transitions maintain visual flow, whereas abrupt changes disrupt continuity and narrative coherence. By enabling fluid transitions between diverse visual vocabularies, video style morphing enriches storytelling depth and expands artistic expression, providing creators with powerful tools to transcend conventional visual storytelling limitations.

Existing video stylization approaches exhibit significant limitations. AnyV2V~\cite{ku2024anyv2v} demonstrates a tendency to overfit style image features to individual frames in temporal sequences. While StyleID~\cite{chung2024style} advances the field by adapting image editing techniques~\cite{tumanyan2023plug, cao2023masactrl} to stylization through self-attention injection with impressive results, these methods—derived primarily from image-based style transfer paradigms~\cite{tumanyan2023plug, cao2023masactrl, yu2024anyedit}—encounter three fundamental challenges when applied to video style morphing:
(1) \textbf{Structural Consistency:} Maintaining coherent object structures across sequential frames despite motion and temporal dynamics remains problematic, as evidenced by layout distortions shown in Fig.~\ref{fig:intro_case} (left).
(2) \textbf{Style Transition:} Achieving gradual, perceptually natural style evolution over time presents difficulties, with existing methods often producing jarring visual discontinuities during style transitions (Fig.~\ref{fig:intro_case}, center).
(3) \textbf{Content Preservation:} Retaining the integrity of source content during stylistic transformation proves challenging, frequently resulting in loss of important visual details, particularly in background elements (Fig.~\ref{fig:intro_case}, right).
These limitations stem largely from the underlying attention mechanisms that are optimized for single reference style images rather than continuous style transformations across temporal sequences.

In this work, we propose SOYO (Style-mOrphing with Adaptive sYnergy for AttentiOn Injection), an end-to-end, open-domain framework for video style morphing that enables smooth style transitions without fine-tuning. Building upon style injection techniques, we enhance the Stable Diffusion framework by strategically manipulating self-attention mechanisms during the denoising process, effectively preserving structural content from source video frames while seamlessly integrating textural elements from style images. As illustrated in Fig.~\ref{fig:method}, our approach comprises four key components:
(1) \textbf{Structural Preservation through Query Injection:} We inject Query features from the original video inversion during denoising to maintain structural integrity of the source content.
(2) \textbf{Progressive Interpolation for Texture Variation:} Key and Value features from the inversion of both style images are dynamically interpolated with time-varying weights during denoising, facilitating smooth textural transitions.
(3) \textbf{AdaIN for Style Modulation:} Statistical parameters (mean and variance) extracted from both style references are progressively blended using temporally-adaptive weights, ensuring gradual and coherent transitions in color distribution and textural characteristics.
(4) \textbf{Adaptive Sampling for Smooth Transitions:} We implement a perceptually-guided schedule that dynamically adjusts blending weights based on style distance metrics, producing temporally balanced transitions that avoid visual discontinuities.

% Specifically, during the denoising process of stylized video frames, we inject the Query from the original video inversion process to preserve the structural information of the original video. For the two stylized images, we save the Key and Value from their inversion process and interpolate them with increasing weights during the denoising of the stylized video, which is then injected to achieve texture variation. 
% To transfer the color and tonal characteristics of the style images to the video frames, we employ AdaIN to modulate the latent representations of the video frames. We derive statistical style parameters (mean and variance) from the inverted latents of both style references. These parameters are progressively blended using time-varying weights to generate intermediate-style representations, which are then applied to the video latents via AdaIN. This ensures a gradual shift in color distribution and texture across frames while preserving structural coherence. To achieve smooth and balanced transitions between styles, we further introduce an adaptive sampling schedule that dynamically adjusts the blending weights based on perceptual style distances, avoiding abrupt changes and ensuring temporally consistent morphing effects. 

To the best of our knowledge, this is the first implementation of video style morphing using diffusion models. Unlike traditional methods, SOYO preserves the precise structure of video frames. Compared to previous diffusion-based methods, SOYO enables smooth style transitions and can handle open-domain style images as shown in Fig.~1. SOYO effectively handles various style combinations across different scenes without requiring specific training. We hope that our work will expand the scope of video stylization tasks, achieving visual effects that were previously unattainable with a single style, through the seamless switching and transitioning between multiple specific styles. In summary, our main contributions are as follows:
\begin{itemize}
    \item We propose a cross-frame style-attention fusion mechanism that maintains structural coherence and content integrity while ensuring seamless texture transitions through dual-style interpolation. 
    % To tackle the challenge of consistent color and luminance shifts, the Dual-Style AdaIN interpolates style parameters, while an adaptive scheduling strategy balances transitions between divergent styles, addressing imbalances in style morphing.
    \item We introduce a multi-scene benchmark for video style morphing, featuring diverse video sequences paired with a broad spectrum of artistic styles. This dataset facilitates comprehensive evaluation of video style morphing to accommodate broader requirements of artistic production.
    \item Extensive experiments demonstrate that our framework outperforms state-of-the-art methods across diverse content-style combinations.
\end{itemize}

\section{Related Work}

\subsection{Image Style Transfer}
Image style transfer has been extensively studied in recent years\cite{zhu2017unpaired, kwon2022clipstyler, huang2024diffstyler}. It transfers the style of one image to a content image while preserving the structure of the original content image. \cite{tgatys2016image} propose a style transfer method based on CNNs, utilizing a pre-trained VGG network to extract both content and style features. Building upon this, \cite{johnson2016perceptual} achieves real-time style transfer by employing perceptual loss. More recently, Adaptive Instance Normalization (AdaIN)\cite{huang2017arbitrary} significantly enhances the speed and flexibility of style transfer by aligning the mean and variance of feature maps with those of the style features. In the realm of diffusion models, StyleDiffusion \cite{wang2023stylediffusion} achieves the decoupling of style and content by explicitly extracting content information while implicitly learning style features. Furthermore, Style Injection in Diffusion \cite{chung2024style} accomplishes style transfer by manipulating the self-attention mechanisms within diffusion models.

\subsection{Video Style Transfer}
Directly applying image stylization models to video stylization tasks may lead to temporal inconsistency issues. Early approaches enhance video consistency by leveraging optical flow \cite{gao2020fast, huang2017real}. Diffusion models have demonstrated remarkable potential in zero-shot video style transfer. AnyV2V~\cite{ku2024anyv2v} edits first frame with a stylization model and performs temporal feature injection on the diffusion model to achieve the entire video stylization. However, this approach tends to cause overfitting to the style image. Style-A-Video \cite{huang2024style} employs a generative pre-trained transformer combined with an image latent diffusion model to achieve concise text-controlled video stylization. Diffutoon~\cite{duan2024diffutoon} addresses the challenges in video stylization by decomposing the video style coloring problem into four sub-problems, while BIVDiff \cite{shi2024bivdiff} uses a decoupled framework with an image diffusion model for frame-by-frame video generation and applies video smoothing after mix inversion. These methods rely on textual conditioning for stylization, making it difficult to accomplish open-domain, style-image-based video style morphing.
\section{Method}

\begin{figure*}[h]
    \centering
    \includegraphics[width=1\linewidth]{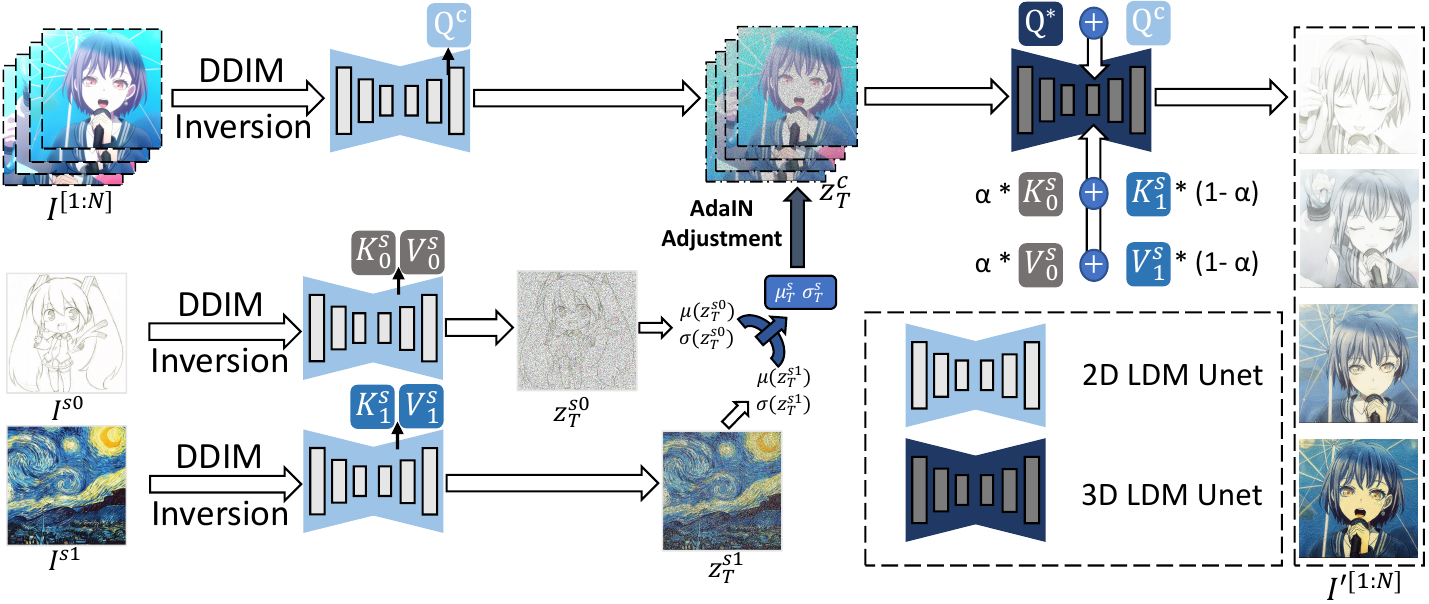} 
    \caption{SOYO pipeline. We perform DDIM Inversion on the source video $I^{[1:N]}$ to obtain $z^c_T$, and execute DDIM Inversion on two style images $I^{s0}$ and $I^{s1}$ to get $z^{s0}_T$ and $z^{s1}_T$, respectively, while saving their attention values.Subsequently, we perform linear interpolation on \( z^{s0}_t \) and \( z^{s1}_t \) with specific weights to obtain \( z^s_t \), which is then used to modulate \( z^c_t \) via AdaIN. During the denoising process, the latents corresponding to each frame are injected with interpolated $K^s$ and $V^s$ values from the style images, while receiving $Q^c$ injections from the source video frames. This results in smoothly transitioned stylized video frames $I'^{[1:N]}$.} 
    \label{fig:method} 
\end{figure*}

To address this challenging problem, we propose SOYO, a novel approach that leverages a pre-trained Text-to-Image (T2I) Stable Diffusion (SD) model to enable tuning-free Video Style Morphing. For simplicity, the subsequent discussion omits the details of the encoding and decoding processes of the latent diffusion model (LDM) autoencoder. As illustrated in Fig.~\ref{fig:method}, our pipeline consists of three key parts: 1) We first perform DDIM inversion on both the source video and style images to extract their latent representations and attention features. 2) We then adapt style characteristics by interpolating statistical features between the style references and modulating the video content via AdaIN. 3) During denoising, we preserve structural information from the source video while seamlessly blending stylistic elements from both reference images, enabling smooth transitions between distinct artistic styles.

\subsection{Preliminaries}
\textbf{Video Style Morphing.} For video style morphing, the task is to generate a video $I'^{[1:N]}$ that maintains the structural consistency of a given input video $I^{[1:N]}$ while smoothly transitioning its visual style from the style of an image $I^{s0}$ to that of another image $I^{s1}$ over time. Different from single image style transfer, this is non-trivial as it requires maintaining temporal consistency of the video content and ensuring smooth transitions between multiple styles.

\noindent\textbf{Probabilistic Diffusion models.} 
Diffusion models~\cite{ho2020denoising, song2020denoising} are probabilistic models designed to generate data samples through a sequence of denoising autoencoders~$\epsilon_{\theta}(x_t, t); t = 1...T$ that estimate the score of a data distribution. This process iteratively adds Gaussian noise \(\epsilon \sim \mathcal{N}(0, I)\) to a clean sample~\(z_0\) according to a variance schedule \(\beta_1, \dots, \beta_t\):
\begin{equation}
    q(z_t|z_{t-1}):=\mathcal{N}(z_t;\sqrt{1-\beta_t}z_{t-1}, \beta_tI), t = 1, 2, \dots, T,
\end{equation}

Our method builds on the Stable Diffusion (SD) model~\cite{rombach2022high}, which performs denoising in a compact latent space to enhance efficiency. The pipeline first maps input images into this latent space using a VAE encoder~\cite{kingma2013auto}, applies the diffusion process and then decodes the refined latent representation back into the image space. In the noise-predicting network $\epsilon_{\theta}$, residual blocks generate intermediate features $f_t^l$, which are further processed using attention mechanisms. Specifically, self-attention captures long-range spatial dependencies within the features:
\begin{equation}
    Attention(Q, K, V) = \text{softmax}\left(\frac{QK^\top}{\sqrt{d_k}}\right) \cdot V
\end{equation}
where $Q$, $K$, and $V$ represent queries, keys, and values derived from the same feature map, with $d_k$ indicating the key/query dimension; cross-attention integrates the textual prompt $P$ by using it to generate keys and values, thereby merging the semantics of both text and image. These attention layers in the SD model play a key role in guiding image composition and synthesis, enabling edits by manipulating attention during the denoising process.

\subsection{Cross-Frame Style Fusion Attention}
To ensure both temporal consistency and spatial coherence across the generated video frames, we extend the Latent Diffusion Model (LDM) into 3D space. In this extension, we modify the self-attention mechanism to incorporate cross-frame attention, which is defined as follows:

\begin{equation}
\begin{aligned}
Q^i &= W^Q z^i, \\
K^i &= W^K [z^{i-1}, z^i, z^{i+1}], \\
V^i &= W^V [z^{i-1}, z^i, z^{i+1}].
\end{aligned}
\end{equation}

To mitigate overfitting to the style image, it is essential to ensure that stylized video frames derive structural information from the source video while acquiring textural details from the style image. Prior research has demonstrated that query (Q) components predominantly determine structural elements, while keys (K) and values (V) significantly impact texture and style features. Our approach leverages this property by strategically decoupling structure preservation from style transition.

We begin by performing DDIM inversion on the source video to extract its self-attention features $Q^c$ along with the noise term $z_T^c$, and similarly extract features $\{K^s_0, V^s_0\}$, $\{K^s_1, V^s_1\}$ and noise terms $z_T^{s_0}$, $z_T^{s_1}$ from the two style images. These features guide the generation of stylized video frames during the denoising process.

The query features play a critical role in capturing structural characteristics of the generated frames. We continuously incorporate $Q^c$ from the source video by setting $Q_t = Q^c_t$ across all timesteps, ensuring consistent preservation of structural elements while managing textural aspects through the key and value features.

\begin{figure}[h]
    \centering
    \includegraphics[width=1\linewidth]{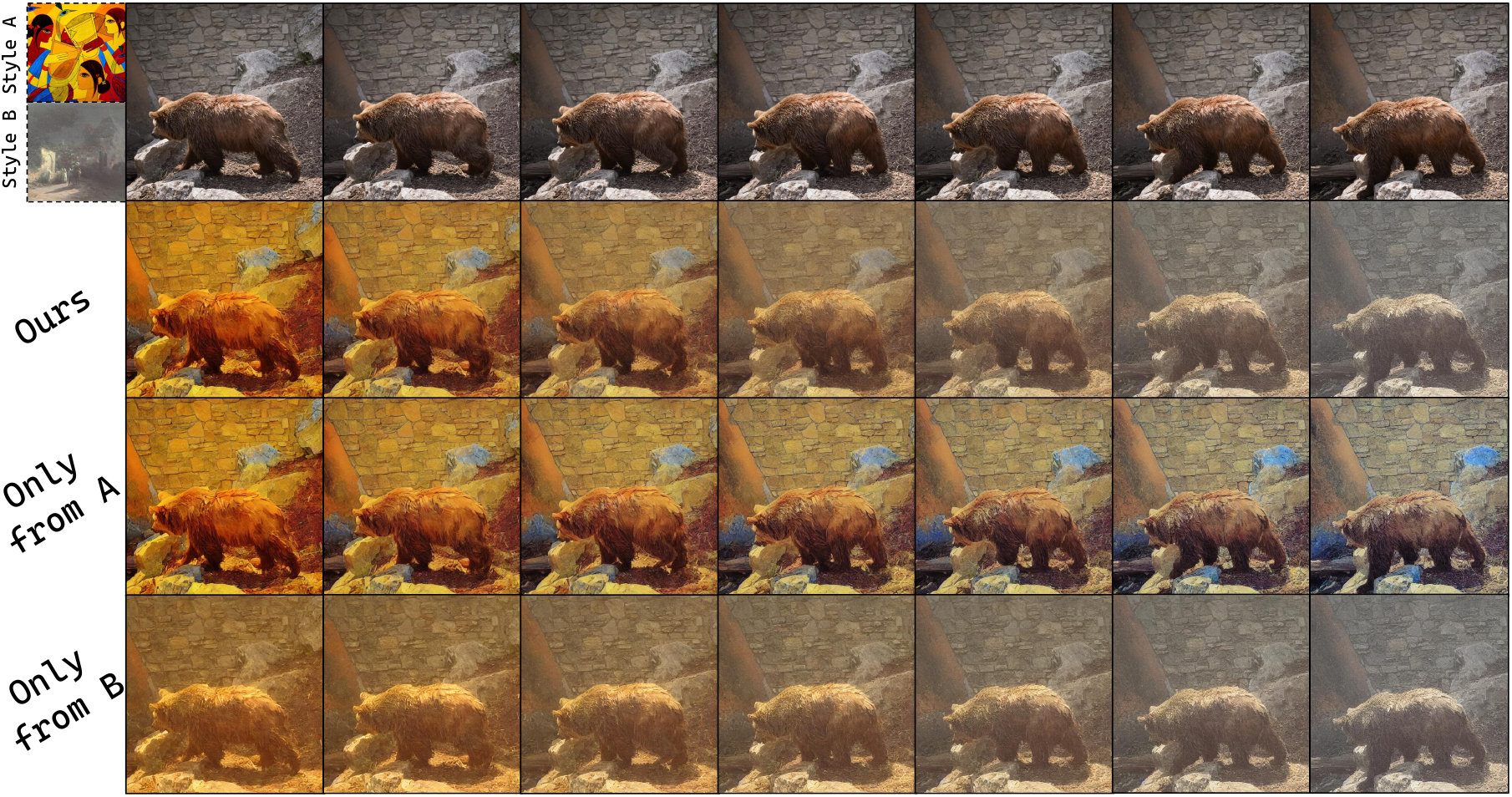} 
    \caption{The injection of keys and values derived from a single style image results in stylized video frames that fail to accurately represent the desired textures and transitions.} 
    \label{fig:injection} 
\end{figure}

In contrast to previous approaches that inject self-attention features from a single style image, our method generates intermediate transitional states by adaptively interpolating both key and value features from two distinct style images, as illustrated in Fig. \ref{fig:injection}:

\begin{equation}
\begin{aligned}
K^m_t &= \text{LinearInterp}(K^s_{0,t}, K^s_{1,t}, \alpha), \\
V^m_t &= \text{LinearInterp}(V^s_{0,t}, V^s_{1,t}, \alpha),
\end{aligned}
\end{equation}

where $\alpha$ represents interpolation coefficients controlling the gradual transition between styles. The modified self-attention process is calculated as:

\begin{equation}
    Attention(Q^c_t, K^m_t, V^m_t) = \text{softmax}\left(\frac{{Q^c_t(K^m_t)}^\top}{\sqrt{d_k}}\right)\cdot V^m_t
\end{equation}

This approach preserves the structural integrity of the source video while gradually infusing style through interpolated $K$ and $V$ features, leading to temporally coherent and stylistically consistent results. As shown in Fig. \ref{fig:injection}, using keys and values from only a single style image cannot achieve smooth transitions between styles, underscoring the necessity of our dual-style approach.

We implement this attention injection mechanism in the last five decoder layers of the diffusion model, where high-level structural and textural features are processed. The complete pipeline is illustrated in Fig. \ref{fig:method}, demonstrating how our method integrates structure preservation and smooth style transition through strategic attention injection.

\subsection{Dual-Style Latent AdaIN Modulation}

While attention mechanisms effectively control structural and textural elements, they alone cannot ensure consistent color tone transitions essential for smooth video style morphing. To ensure this, we employ the Adaptive Instance Normalization (AdaIN)~\cite{huang2017arbitrary} technique, which adjusts the latent representations of the target frames by modulating them with the style statistics (mean and standard deviation) extracted from the reference images.

In video style morphing, where the objective is to gradually transition the visual style of a video from one reference image to another, we interpolate between the mean and standard deviation of the latents derived from the two reference images. This interpolation is governed by a time-dependent parameter \( \alpha \), which adaptively controls the degree of style blending between the two reference styles across the video frames, ensuring a smooth stylistic transition.

Let \( z_t^c \) denote the latent features of the target video frame, and let \( z_t^{s0} \) and \( z_t^{s1} \) represent the latents derived from the two reference images. The first step is to compute the mean \( \mu \) and standard deviation \( \sigma \) of these latents. The interpolated mean \( \mu_t^s \) and standard deviation \( \sigma_t^s \) for the style transfer are then computed as a weighted average between the two reference latents:

\begin{equation}
    \mu_t^s = \alpha \cdot \mu(z_t^{s0}) + (1 - \alpha) \cdot \mu(z_t^{s1})
\end{equation}

\begin{equation}
    \sigma_t^s = \alpha \cdot \sigma(z_t^{s0}) + (1 - \alpha) \cdot \sigma(z_t^{s1})
\end{equation}

Here, \( \alpha \) is the interpolation coefficient that transitions from 0 (first reference image style) to 1 (second reference image style) as the video progresses. 

The AdaIN modulation is applied to the video frame latents \( z_t^c \) using the interpolated style statistics \( \mu_t^s \) and \( \sigma_t^s \):

\begin{equation}
    z_t = \sigma_t^s \left( \frac{z_t^c - \mu(z_t^c)}{\sigma(z_t^c) + \epsilon} \right) + \mu_t^s
\end{equation}

Where \( \mu(z_t^c) \) and \( \sigma(z_t^c) \) represent the mean and standard deviation of the source video frame latent \( z_t^c \), and \( \epsilon \) is a small constant added to avoid division by zero. This operation ensures that the generated video frame \( z_t \) is stylized with the desired visual features while preserving the structural consistency of the original content.

By progressively modulating the interpolation coefficient \( \alpha \) throughout the video sequence, we achieve a seamless transition between the two reference styles, ensuring both temporal coherence and stylistic fidelity. This approach enables a smooth and continuous evolution of style across video frames, preserving the structural integrity of the target content while adapting its color and texture characteristics to align with the desired reference styles.

\subsection{Adaptive Style Distance Mapping}
To address the issue of imbalanced style transformation in video style transfer, where specific style pairs lead to asymmetric transitions—resulting in a slower style change on one side and more than half of the frames being biased toward a particular style—we propose a balanced control method, Adaptive Style Distance Mapping (ASDM).

First, we perform pre-sampling from the 10th denoising step to rapidly generate stylized frames. These frames are then processed through a pre-trained VGG16 network~\cite{simonyan2014very} to compute style distances, measured by differences in mean and standard deviation across feature layers between the generated frames and both reference style images. Based on these distance curves, we identify the perceptual style transition point, denoted as $\alpha_{\text{mid}}$, which represents the normalized frame index where the style distance curves intersect. When $\alpha_{\text{mid}} \neq 0.5$, this indicates an imbalanced style transition that requires compensation.

Next, we construct a linear transition curve for $\alpha$ using $\texttt{LinSpace}(0,1)$ over frames. To emphasize regions with more pronounced perceptual style changes, we introduce an adaptive scaling factor that modulates the sampling density:

\begin{equation}
S_{\text{max}} = 1 + \lambda_{\alpha} \, \cdot|\alpha_{\text{mid}} - 0.5|
\end{equation}

where $\lambda_{\alpha}$ is a hyperparameter controlling the intensity of the compensation (default value $\lambda_{\alpha}=5$). The relationship between $\alpha_{\text{mid}}$ and $S_{\text{max}}$ is proportional to the deviation from the balanced midpoint: the greater the imbalance, the larger the $S_{\text{max}}$ value, resulting in stronger compensation.

If $\alpha_{\text{mid}} < 0.5$, the scaling factor gradually increases from 1 to $S_{\text{max}}$ across frames; otherwise, it decreases from $S_{\text{max}}$ back to 1. These scaling factors modulate the differences between consecutive values in the baseline curve, and their cumulative sum—normalized to the range $[0,1]$—produces a non-uniform $\alpha$ curve with higher sampling density in regions of significant perceptual style change.

While this adaptive approach introduces approximately 20\% additional computational overhead due to the pre-sampling step, the cost is acceptable given our method's overall efficiency. With our implementation on an RTX 3090 GPU, the DDIM inversion requires less than 5 seconds, and the complete sampling process takes less than 15 seconds, making the additional computational cost for achieving balanced style transitions well justified for high-quality video style morphing.
\section{Experiments}
In this section, we first introduce an novel SOYO-Test benchmark with various scenrios for challenging video style morphing evaluation (§~\ref{exp4.2}). 
We then present both quantitative~(§~\ref{exp4.3}) and qualitative~(§~\ref{exp4.4}) results and corresponding analyses to assess SOYO’s superiority on addressing the video style morphing in both temporal consistency and style fidelity. Finally, we conduct in-depth analysis (§~\ref{exp4.5}) to explore the contribution of each component.
% We then comprehensively compare SOYO with state-of-the-art approaches on this benchmark (§~\ref{exp4.3}), demonstrating our method's superior ability to maintain structural consistency while achieving smooth style transitions across diverse artistic styles. We further provide qualitative comparisons (§~\ref{exp4.4}) to visually illustrate the advantages of our approach over existing methods. Finally, we conduct detailed ablation studies (§~\ref{exp4.5}) to analyze the contribution of each component in our framework, confirming that our design choices effectively address the challenges of video style morphing in both temporal consistency and style fidelity.

\subsection{Experimental Setup}\label{exp4.1}
\noindent\textbf{Implementation Details.} In our experiments, we use Stable Diffusion v1.5\cite{rombach2022high} as our base model. During the inversion and denoising process, we adopt the DDIM schedule with 50 steps. Attention injection is applied to the last five layers of the decoder and across all time steps. Since we perform style transfer guided by style images, we do not use classifier-free guidance during either the inversion or denoising process. All experiments are conducted on an RTX 3090 GPU. 

\noindent\textbf{Baselines.} We use the following baselines: (1) \textbf{Traditional methods}, including EFDM\cite{zhang2022exact} and AdaIN\cite{huang2017arbitrary}, take style image pairs as input for image style transfer, requiring feature interpolation to adapt for video style morphing.
(2) \textbf{Diffusion-based methods}: it includes AnyV2V\cite{ku2024anyv2v} that processes videos in two segments due to its single-frame propagation mechanism; Diffutoon\cite{duan2024diffutoon}, which requires text prompts generated by GPT-4o\cite{achiam2023gpt} and uses text embedding interpolation for style transfer; and StyleID\cite{chung2024style}, which employs Dual-Style AdaIN for video style morphing. 
% and two traditional methods (EFDM\cite{zhang2022exact}, AdaIN\cite{huang2017arbitrary}). While most methods use style image pairs, Diffutoon requires text prompts generated by GPT-4o\cite{achiam2023gpt}. Each method is adapted for video style morphing: traditional methods use feature interpolation; diffusion methods employ either Dual-Style AdaIN (StyleID) or text embedding interpolation (Diffutoon); AnyV2V processes videos in two segments due to its single-frame propagation approach. 
More details are in Appendix A.1.

\noindent\textbf{Evaluation Metrics}
We evaluate our method using five metrics to assess perceptual quality, temporal consistency, and style fidelity. LPIPS\cite{zhang2018unreasonable} measures structural preservation between generated and original video frames. CLIP Score\cite{radford2021learning} quantifies semantic consistency across frames by computing cosine similarity between frame embeddings. Perceptual path length (PPL)\cite{karras2020analyzing} calculates cumulative feature differences between adjacent frames, reflecting temporal smoothness. Perceptual distance variance (PDV) measures the standard deviation of these differences, indicating transition stability. StyleLoss evaluates style adherence by comparing Gram matrices of generated frames and reference styles using weighted MSE between frame Gram features and style images, with weights interpolated over time.    

\subsection{SOYO-Test Benchmark}\label{exp4.2}
We introduce the SOYO-Test, a benchmark designed to evaluate dynamic style transitions in videos. The SOYO-Test serves as the first dedicated testbed for evaluating dynamic style transitions in video, pushing the boundaries of existing methods through its rigorous combination of artistic challenges. We detail the benchmark in terms of \textit{data construction}, \textit{scenario diversity}, and \textit{evaluation protocol}.

\noindent\textbf{Data Construction.}
The dataset consists of 25 video sequences from the DAVIS2017 dataset~\cite{pont20172017}, capturing various real-world motions such as animal movements, human activities, and vehicle dynamics. These video clips are paired with 40 distinct artistic styles from the Style Transfer Dataset~\cite{kitov2024style}, which span art movements from the Renaissance to Modernism. The selected styles exhibit diverse chromatic characteristics, ranging from vibrant Fauvist palettes to monochromatic sketches. All videos are standardized to $512 \times 512$ through cropping and resampling.

\noindent\textbf{Various Scenarios.}
To create challenging morphing scenarios, we pair styles with maximal perceptual distance based on three criteria: (1) Contrasting art movements (e.g., Classical Realism vs. Surrealism), (2) Opposing color schemes (e.g., vivid vs. grayscale), and (3) Historical-temporal disparity (e.g., 18th-century portraits vs. modern street art). These pairings result in 20 distinct style combinations, ensuring significant variation in artistic contrasts.

\noindent\textbf{Task and Evaluation.}
The evaluation protocol generates 500 style morphing sequences, allowing for a comprehensive assessment of temporal consistency, structural preservation, and smooth style interpolation. By testing under these controlled yet highly diverse artistic variations, this benchmark enables the systematic evaluation of video style morphing methods.

% We introduce the SOYO-Test with curated content videos and diverse artistic styles for evaluating dynamic style transitions. Our benchmark comprises 25 video sequences from DAVIS2017~\cite{pont20172017} capturing various real-world motions (animal movements, human activities, vehicle dynamics) paired with 40 distinct artistic styles from the Style Transfer Dataset~\cite{kitov2024style}. These styles span major art movements from Renaissance to Modernism and feature diverse chromatic characteristics from vibrant Fauvist palettes to monochromatic sketches. We create challenging morphing scenarios by pairing styles with maximal perceptual distance based on three criteria: 1) Contrasting art movements (e.g., Classical Realism vs. Surrealism), 2) Opposing color schemes (vivid vs. grayscale), and 3) Historical-temporal disparity (18th-century portraits vs. modern street art). This approach yields 20 distinct style pairs. All videos are standardized to $512 \times 512$ resolution through cropping and resampling. Our evaluation protocol generates 500 style morphing sequences, enabling comprehensive assessment of temporal consistency, structural preservation, and smooth style interpolation. This benchmark serves as the first dedicated testbed for evaluating video style morphing methods under controlled yet challenging artistic variations.

\begin{figure*}[ht]
    \centering
    \includegraphics[width=1\linewidth]{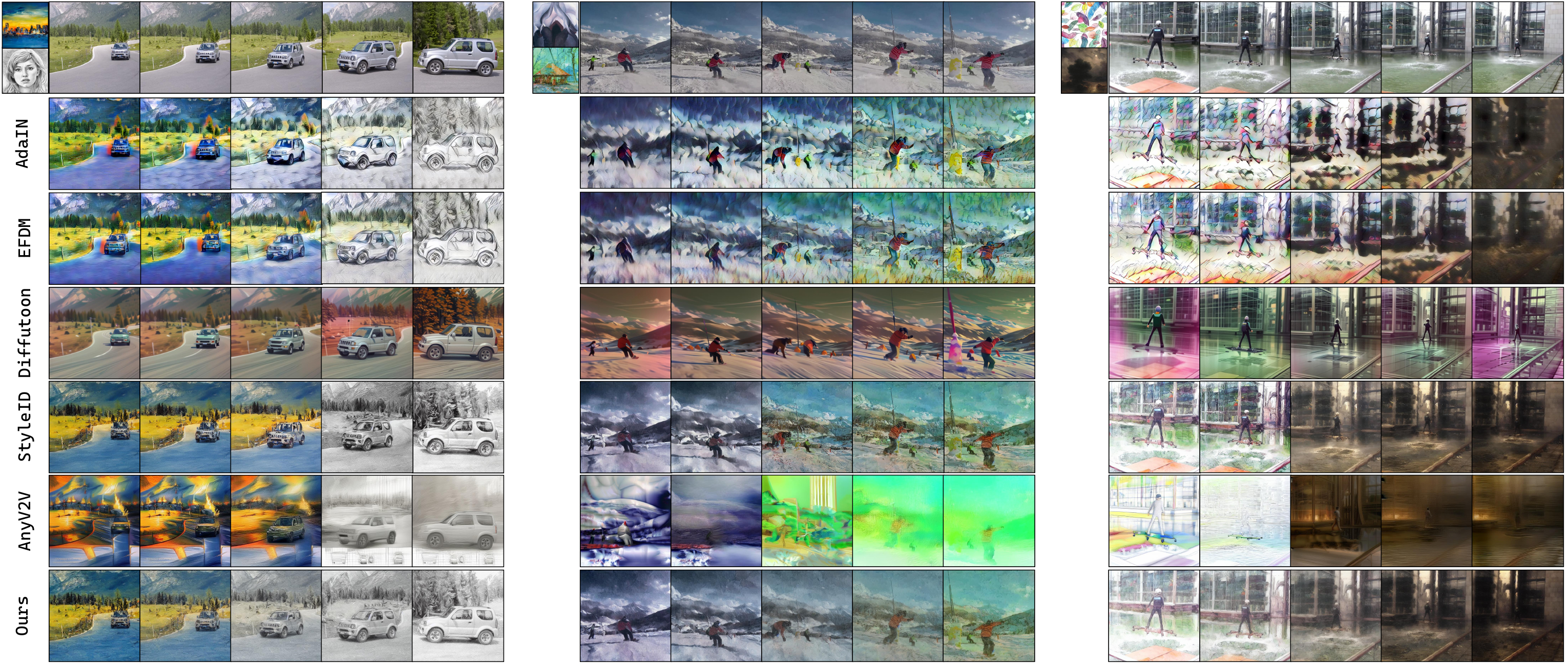} 
    \caption{Qualitative comparisons of SOYO with existing methods.} 
    \label{fig:cmp} 
\end{figure*}

\subsection{Comparison on SOYO-Test Benchmark}\label{exp4.3}
Table~\ref{tab:comparison} presents the results of the SOYO-test benchmark, which includes various video scenes with two distinctly different style images to rigorously evaluate SOYO's superiority across a broader range of challenging style morphing scenarios. Based on the quantitative experiment results, we have summarized the following conclusions:
% \subsection{Quantitative Comparison}
% To evaluate the superiority of our method, we compare it with five state-of-the-art approaches, including three diffusion-based methods, AnyV2V\cite{ku2024anyv2v}, Diffutoon\cite{duan2024diffutoon} and StyleID\cite{chung2024style}, as well as two traditional methods, EFDM\cite{zhang2022exact} and AdaIN\cite{huang2017arbitrary}. All methods use the same style image pair as input, except for Diffutoon, which relies on text prompts. To ensure a fair comparison, we use GPT-4o\cite{achiam2023gpt} to generate textual descriptions of the style images as text-prompt for Diffutoon. We adopt the official publicly available implementations of these methods and make specific modifications tailored for video style morphing. For traditional methods, we interpolate and fuse the features derived from two style images; for diffusionbased methods, we adopt the Dual-Style AdaIN consistent with ours for StyleID, and we apply text embedding interpolation for Diffutoon. Given that AnyV2V is based on single-frame editing propagation, we divide it into two segments to generate stylized video frames separately. 

As shown in Table~\ref{tab:comparison}, our method demonstrates several key advantages: \textbf{\textit{(i)} SOYO achieves superior temporal consistency}, evidenced by the best PPL (116.35 vs. second-best 124.64) and PDV scores (2.35 vs. second-best 3.20), indicating significantly smoother transitions between frames. \textbf{\textit{(ii)} SOYO better preserves structural integrity} of source content, with the lowest LPIPS score (0.248 compared to StyleID's 0.289), confirming that SOYO maintains original video structures rather than imposing style image features. \textbf{\textit{(iii)} The trade-off between structure preservation and style transfer} is revealed in StyleLoss metrics, where traditional methods like EFDM (0.099) outperform ours (0.167) by overfitting to style features at the expense of structural coherence. Visual inspection of these results (Fig.~\ref{fig:cmp}) further confirms this observation. This balanced approach demonstrates SOYO's effectiveness in maintaining the fundamental temporal and structural properties essential for high-quality video style morphing, particularly in challenging scenarios with dramatic style transitions.

\begin{table}[ht]
\centering
\resizebox{\columnwidth}{!}{
\begin{tabular}{|l|c|c|c|c|c|}
\hline
\textbf{Method} & \textbf{LPIPS} $\downarrow$ & \textbf{CLIP Score} $\uparrow$ & \textbf{PPL} $\downarrow$ & \textbf{PDV} $\downarrow$ & \textbf{StyleLoss} $\downarrow$ \\
\hline
AdaIN & 0.4044 & 0.9241 & 238.9440 & 4.1549 & \underline{0.1293} \\
EFDM & 0.3954 & \underline{0.9297} & 194.9744 & 3.2368 & \textbf{0.0987} \\
\hline
Diffutoon & 0.3608 & \textbf{0.9378} & 173.7948 & 3.4504 & 0.7160 \\
StyleID & \underline{0.2887} & 0.9210 & 164.3515 & \underline{3.1981} & 0.1394 \\
AnyV2V & 0.5006 & 0.9037 & \underline{124.6409} & 4.3531 & 0.5488 \\
\hline
\textbf{Ours} & \textbf{0.2482} & \underline{0.9297} & \textbf{116.3492} & \textbf{2.3507} & 0.1674 \\
\hline
\end{tabular}
}
\caption{Quantitative comparison with state-of-the-art methods. Best results are in \textbf{bold}, second-best are \underline{underlined}.}
\label{tab:comparison}
\end{table}

\subsection{Qualitative Comparison}\label{exp4.4}
As illustrated in Fig.~\ref{fig:cmp}, our method achieves smooth style transitions while maintaining stable structures across video frames. For style transfer applications, both AdaIN and EFDM excessively apply style features at the expense of original structural preservation, resulting in significant artifacts. As illustrated in Fig.~\ref{fig:cmp} (third and last rows), despite achieving better StyleLoss than the overfitting EFDM method, our approach still accomplishes effective style transfer while maintaining superior structural integrity. Diffutoon demonstrates limited capability in handling arbitrary style transfer tasks. Notably, StyleID and AnyV2V fail to generate smooth visual transitions through intermediate style phases due to the absence of attention-based hierarchical style feature fusion. These observations on our SOYO-Test benchmark, with its diverse style pairs and content variations, highlight how existing methods struggle with the simultaneous demands of structural preservation and smooth style adaptation required for effective video style morphing, resulting in significantly lower temporal smoothness metrics compared to our approach.

\subsection{Ablation Study}\label{exp4.5}
\begin{figure}[ht]
    \centering
    \includegraphics[width=1\linewidth]{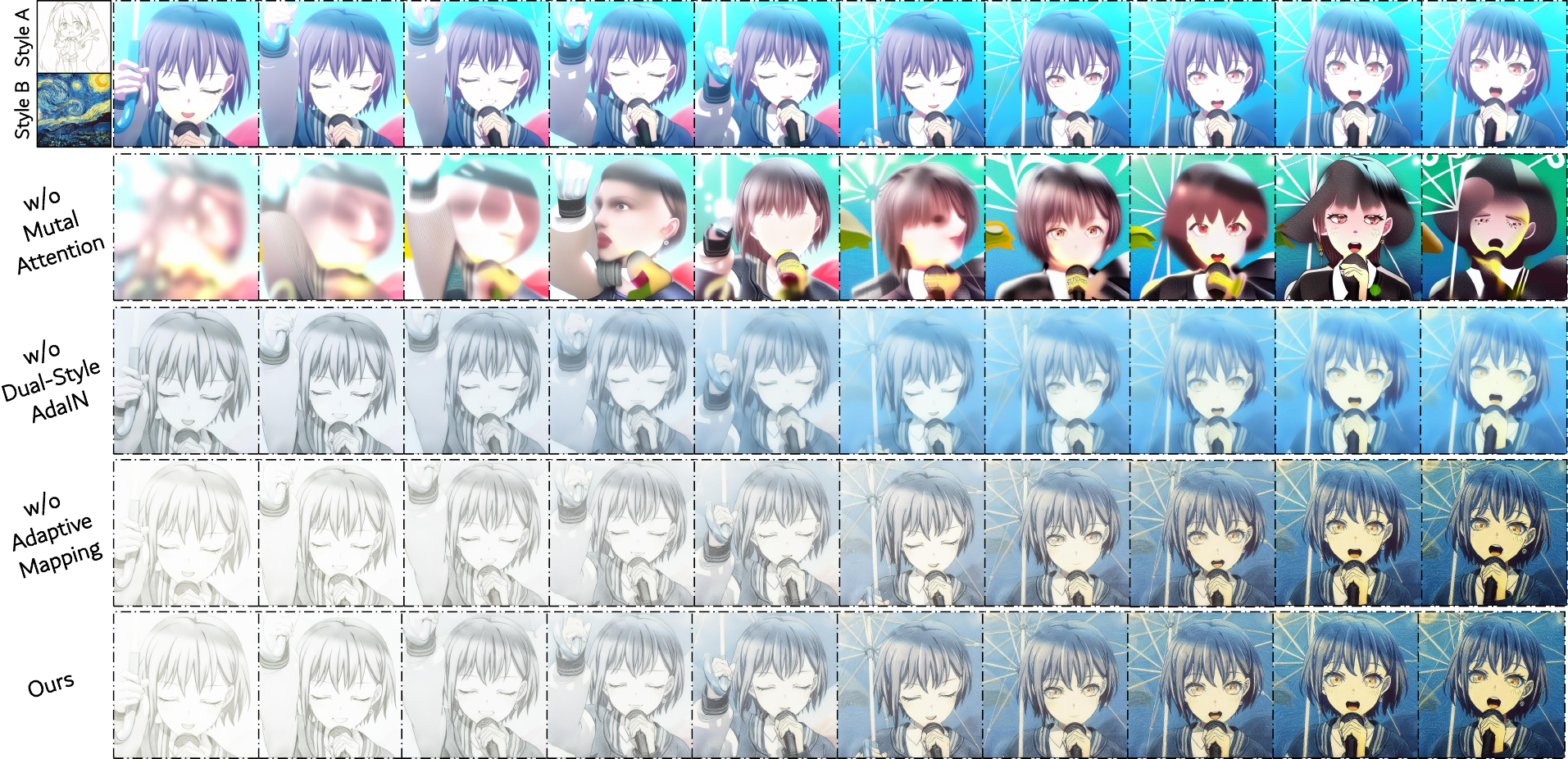} 
    \caption{Qualitative comparison with ablation studies.} 
    \label{fig:ablation} 
\end{figure}

We conduct comprehensive quantitative and qualitative experiments to evaluate the effectiveness of each component in our framework, with results presented in Fig.~\ref{fig:ablation} and Table \ref{tab:ablation}. Mutual attention proves crucial for preserving content structure while enabling smooth texture translation between style images. This component effectively restores structural details that vanilla DDIM inversion fails to capture, while simultaneously incorporating stylistic attributes from target styles, resulting in significant performance improvements (second row, Fig.~\ref{fig:ablation}). The dual-style AdaIN component ensures precise control over color distributions and tonal characteristics, facilitating natural and coherent transitions between different visual styles, as demonstrated in the third row of Fig.~\ref{fig:ablation}. When processing style images with substantial perceptual differences, we observe that unmodified interpolation leads to imbalanced style morphing. Our ASDM effectively addresses this issue by optimizing transition pacing throughout the sequence—mitigating both overly slow initial transitions and excessively rapid later transitions (fourth row, Fig.~\ref{fig:ablation}). While this optimization slightly increases PPL and CLIP scores due to reduced small-change segments, the overall visual quality and transition balance are substantially improved.

\begin{table}[ht]
\centering
\resizebox{\columnwidth}{!}{
\begin{tabular}{|l|c|c|c|c|c|}
\hline
\textbf{Variant} & \textbf{StyleLoss} & \textbf{LPIPS} & \textbf{CLIP Score} & \textbf{PPL} & \textbf{PDV} \\
\hline
w/o Mutal Atten & 0.2681 & 0.3709 & 0.7527 & 200.41 & 3.623 \\
w/o Dual AdaIN & 0.2414 & \textbf{0.1786} & 0.9296 & \textbf{70.72} & \textbf{2.109} \\
w/o ASDM & 0.1679 & 0.2495 & \textbf{0.9297} & 115.69 & 2.336 \\
ours & \textbf{0.1674} & 0.2482 & \textbf{0.9297} & 116.35 & 2.351 \\
\hline
\end{tabular}
}
\caption{Ablation Study Results}
\label{tab:ablation} 
\end{table}

\begin{figure}[h]
    \centering
    \includegraphics[width=1\linewidth]{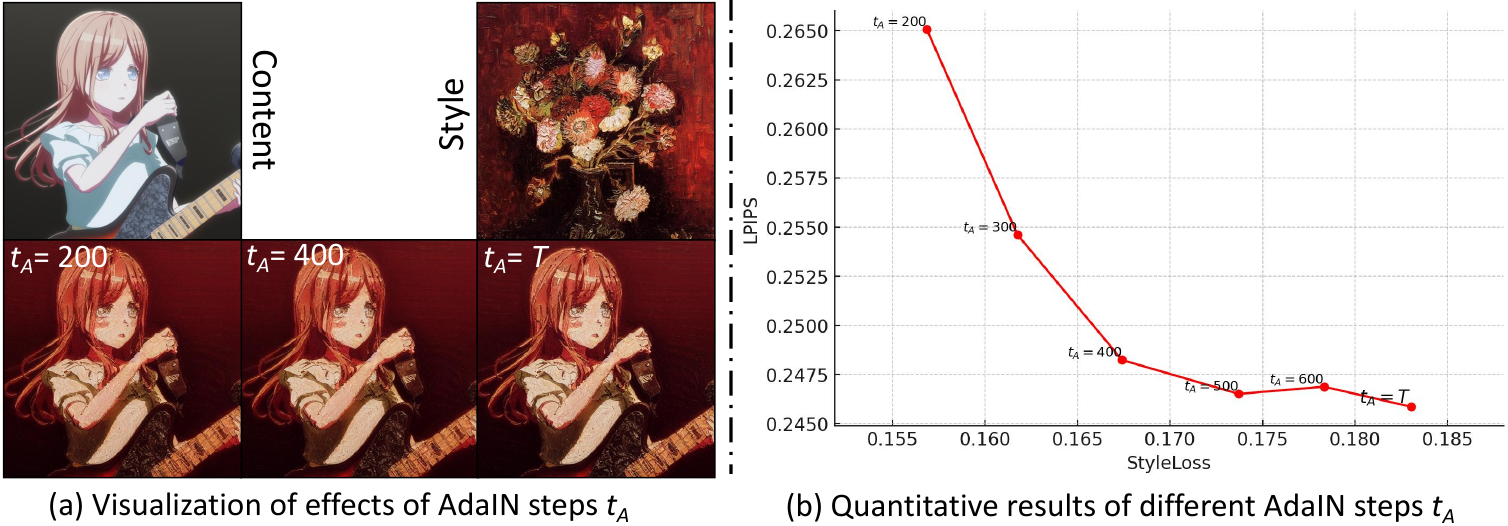} 
    \caption{A smaller \( t_A \) results in greater consistency with the color and brightness of the style image. However, when \( t_A \) is too small, it may compromise the structural integrity of video frames.} 
    \label{fig:ablation_adain} 
\end{figure}

\noindent\textbf{AdaIN steps}
In the denoising phase, we initiate the AdaIN modulation at \( t = T \) and cease modulation after \( t = t_A \). To investigate the influence of the modulation duration, we conduct an ablation study on the Dual Style AdaIN range by varying \( t_A \) and evaluating both LPIPS and Style Loss. The results are presented in Fig.~\ref{fig:ablation_adain}. Notably, when \( t_A = 400 \), an optimal balance is achieved between structural preservation and the degree of stylization in the video frames.

\begin{figure}[h]
    \centering
    \includegraphics[width=1\linewidth]{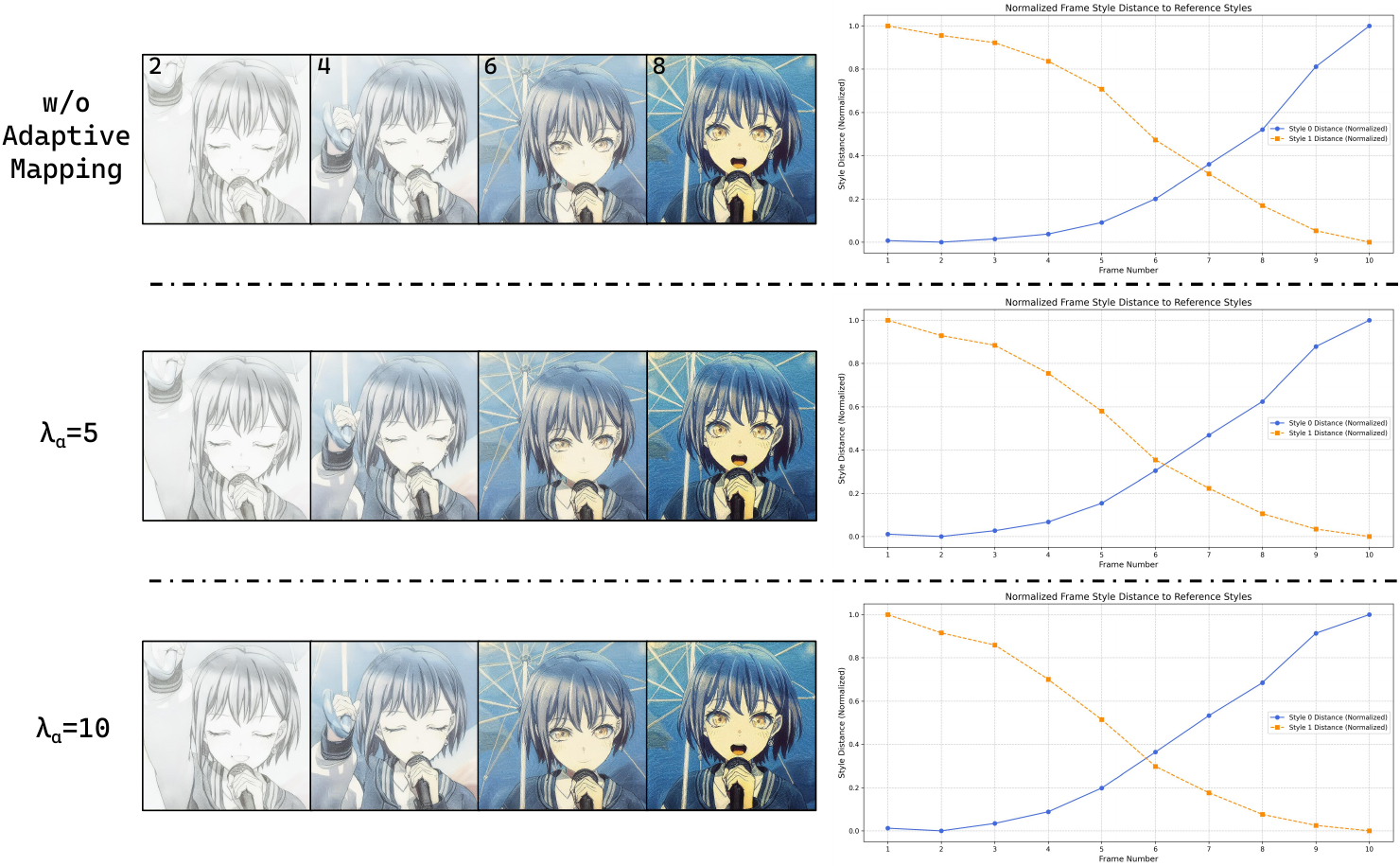} 
    \caption{Comparison with different $\lambda_{\alpha}$.} 
    \label{fig:alpha} 
\end{figure}

\noindent\textbf{Adaptive alpha mapping}
To address imbalanced style morphing, we employ Adaptive Style Distance Mapping (ASDM) to adjust the alpha sequence. As shown in Fig.~\ref{fig:ablation}, our ablation study on $\lambda_{\alpha}$ parameters (visualized in Fig.~\ref{fig:alpha}) reveals that without ASDM, the style transition midpoint exhibits significant shift, causing most of the frames to predominantly reflect Style A. ASDM effectively mitigates this imbalance, particularly when setting $\lambda_{\alpha}=10$. However, we adopt a conservative default $\lambda_{\alpha}=5$ configuration to prevent potential over-correction that might shift the balance point towards the opposite style extreme.

\section{Conclusion}
We present SOYO, a tuning-free framework addressing the open-domain video style morphing challenge through style image-guided synthesis with high fidelity. Our approach enables smooth style transitions by strategically blending interpolated key-value pairs from the style image inversion process into the video frame denoising process, thereby generating intermediate frames with progressive texture transformations. The proposed Dual Style AdaIN mechanism ensures photometric consistency through coordinated luminance and chrominance transitions across frames. To mitigate style transition imbalances, we introduce a pre-sampling-based adaptive weight scheduling algorithm that optimizes interpolation coefficients. For a comprehensive evaluation, we establish the SOYO-Test Benchmark containing diverse style-content pairs. Extensive experiments demonstrate SOYO's superior performance over existing state-of-the-art methods.
%\appendix
%\input{sec/6_appendix}

%\input{sec/7_formatting}
%\input{sec/8_finalcopy}

{
    \small
    \bibliographystyle{ieeenat_fullname}
    \bibliography{main}
}

\end{document}